\documentclass{acmart}

\pdfoutput=1

\usepackage{array}
\usepackage{soul}
\usepackage{academicons}
\usepackage{orcidlink}

\AtBeginDocument{%
  \providecommand\BibTeX{{%
    \normalfont B\kern-0.5em{\scshape i\kern-0.25em b}\kern-0.8em\TeX}}}

\copyrightyear{2021}
\acmYear{2021}
\setcopyright{acmlicensed}\acmConference[ETRA '21 Full Papers]{2021 Symposium on Eye Tracking Research and Applications}{May 25--27, 2021}{Virtual Event, Germany}
\acmBooktitle{2021 Symposium on Eye Tracking Research and Applications (ETRA '21 Full Papers), May 25--27, 2021, Virtual Event, Germany}
\acmPrice{15.00}
\acmDOI{10.1145/3448017.3457380}
\acmISBN{978-1-4503-8344-8/21/05}

\setcitestyle{square}

\begin{document}

\title{Neural Networks for Semantic Gaze Analysis in XR Settings}
\author{Lena Stubbemann\, \orcidlink{0000-0001-8725-7092}}
\orcid{0000-0001-8725-7092}
\affiliation{%
  \department{Department of Quality and Process Management}
  \institution{Department of Quality and Process Management, University of Kassel}
  \city{Kassel}
  \country{Germany}}
\email{stubbemann@uni-kassel.de}
\authornote{Both authors contributed equally to the paper}

\author{Dominik Dürrschnabel\, \orcidlink{0000-0002-0855-4185}}
\orcid{0000-0002-0855-4185}
\affiliation{%
  \department{Knowledge \& Data Engineering Group}
  \institution{Knowledge \& Data Engineering Group, University of Kassel}
  \city{Kassel}
  \country{Germany}}
\additionalaffiliation{%
  \department{Interdisciplinary Research Center for Information System Design, }
  \institution{University of Kassel}
  \city{Kassel}
  \country{Germany}}
\email{duerrschnabel@cs.uni-kassel.de}
\authornotemark[2]

\author{Robert Refflinghaus}
\affiliation{%
  \department{Department of Quality and Process Management}
  \institution{Department of Quality and Process Management, University of Kassel}
  \city{Kassel}
  \country{Germany}}
\email{refflinghaus@uni-kassel.de}

\titlenote{\textcopyright{Stubbemann, L. Dürrschnabel, D., Refflinghaus R.} 2021. This is the author's version of the work. It is posted here for
your personal use. Not for redistribution. The definitive version was published
in ETRA '21 Full Papers, May 25–27, 2021, \url{https://doi.org/10.1145/3448017.3457380}}

\renewcommand{\shortauthors}{Stubbemann, Dürrschnabel, and Refflinghaus}


\begin{abstract}

  Virtual-reality (VR) and augmented-reality (AR) technology is increasingly combined
  with eye-tracking.
  This combination broadens both fields and opens up new areas of application, in which visual perception and related cognitive processes can be studied in interactive but still well controlled settings.
  However, performing a semantic gaze analysis of eye-tracking data from interactive three-dimensional scenes is a resource-intense task, which so far has been an obstacle to economic use.
  In this paper we present a novel approach which minimizes time and information necessary to annotate volumes of interest (VOIs) by using techniques from object recognition.
  To do so, we train convolutional neural networks (CNNs) on synthetic data sets
  derived from virtual models using image augmentation techniques.
  We evaluate our method in real and virtual environments, showing that the method
  can compete with state-of-the-art approaches, while not relying on additional markers or preexisting databases but instead offering cross-platform use.

\end{abstract}

\ccsdesc[500]{Human-centered computing~Virtual reality}
\ccsdesc[500]{Computing methodologies~Object recognition}
\ccsdesc[500]{Computing methodologies~Supervised learning by classification}
\ccsdesc[500]{Computing methodologies~Image processing}
\ccsdesc[500]{Computing methodologies~Perception}

\keywords{Volumes of Interest (VOI), Semantic Gaze Analysis, Synthetic Training Data, Neural Network, User Centered Dynamic Recordings}

\begin{teaserfigure}
\includegraphics[width=\textwidth]{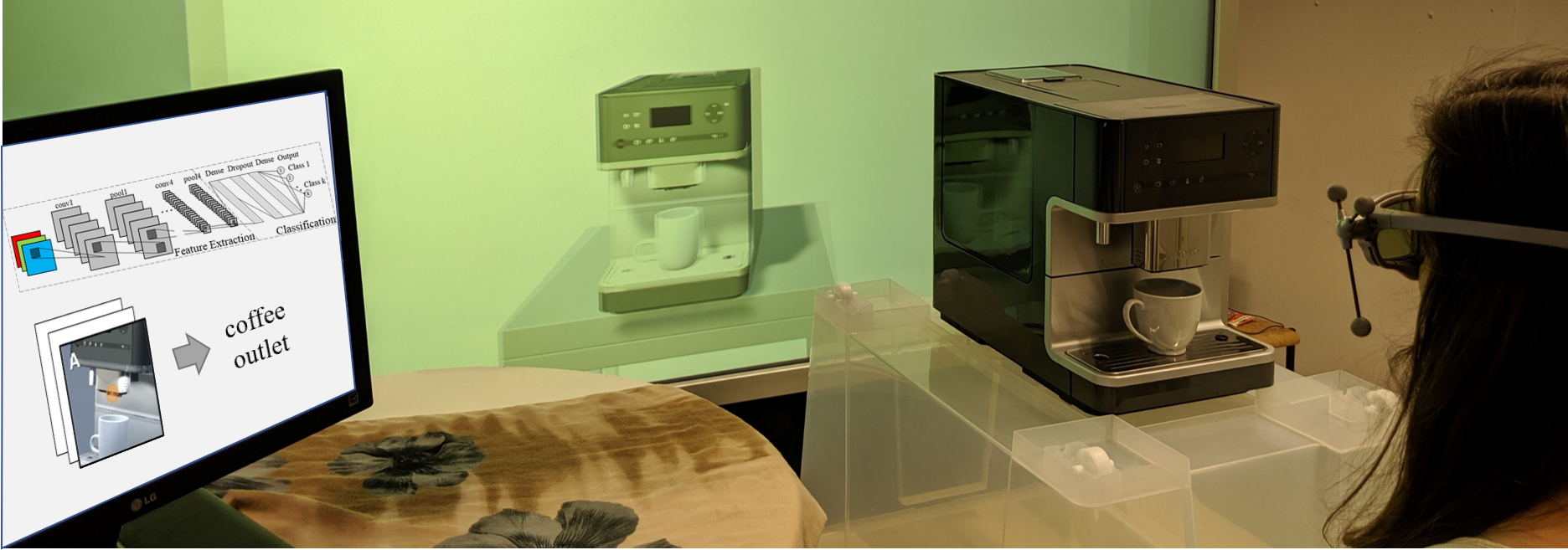}
\caption{Experimental design research (right) is one example among many, where augmented and virtual reality technologies are used in combination with eye-tracking. This work addresses how semantic gaze analysis in real-world and virtual-reality settings can be done using convolutional neural networks (left). \textcopyright{Lena Stubbemann}}
\Description{One can see a person with mobile eye tracking classes looking at a real product and a virtual product displayed stereoscopically on a projection panel. In the left one can see a screen with a neural network analysing an image upon the product feature looked at.}
\end{teaserfigure}

\maketitle

\section{Introduction}
\label{sec:introduction}

Increasing attempts are being made to investigate human perception systematically and as objectively as possible using biometrics such as eye tracking \cite{Borgianni.2020}.
This development goes along with the availability of continuously improved eye-trackers and the increasing use of eye-tracking with XR environments and applications.
Thereby, visual perception and related cognitive processes can be studied in interactive but still well controlled settings.
However, for many use cases
visual patterns and oculometric parameters alone are insufficient to access relevant information about users. Rather, it must additionally be determined what subjects are looking at.
The process to identify which objects or features subjects place their visual and cognitive attention upon is referred to as \emph{semantic gaze analysis}.

In a video frame the two-dimensional depiction of an object is usually refered to as \emph{region of interest}.
Thus, a \emph{volume of interest} is commonly defined by the intersection of three-dimensional extrusions of all regions of interest in different video frames in \cite{Duchowski.1997}.
In this work we therefore refer to \emph{volumes of interest (VOIs)} as the three-dimensional object that emerges from this intersection, which we derive directly from the three-dimensional bodies of the studied object.

What makes the annotation challenging is that interactive three-dimensional scenes result in user-specific gaze videos with constantly changing perspectives on the target object.
VOIs can thereby move, vanish, reappear, and change shape, size or illumination, aggravating the semantic gaze analysis~\cite{Kurzhals.2017}.
As soon as participants freely interact with dynamic three-dimensional stimuli (i.e. prototypes) in virtual or real-world settings, manual annotations are thus still considered a standard procedure ~\cite{Holmqvist.2011}.
However, besides being time-intensive, this method is also prone to evaluator effects~\cite{Kurzhals.2017}.
Even though some approaches try to tackle those problems, available methods still require large resource and time investments (cf.~Section~\ref{sec:related_work}).
When it comes to resources, especially data science approaches for semantic gaze analysis suffer from a lack of suiting annotated training data sets and thus to often fail to be widely applied.
This is particularly true when VOIs need to be annotated on a feature instead of an object level,  e.g. in product design, medical or usability tasks.
Hence, the problem of inefficient annotation is regarded as one of the biggest challenges for the economic use of eye tracking in interactive three-dimensional scenes and is currently rendering many studies unfeasible~\cite{Pfeiffer.2014}.

In this work we present a method which aims to reduce the necessary resources involved in the process of annotating VOIs in interactive three-dimensional scenes.
While other approaches either require additional motion tracking systems~\cite{Paletta.2013,Pfeiffer.2016}, preexisting data sets~\cite{Sattar.2017, Steil.2018} or manually annotated data ~\cite{Toyama.2012,Kurzhals.2015} to address the annotation problem, all we need is a computer aided design (CAD) model or another virtual representation of the scene.
This makes our approach particularly attractive for all disciplines where eye-tracking studies are conducted either on virtual objects, in virtual environments or on real objects with digital twins, which is the case for most use cases in virtual-reality (VR) and augmented-reality (AR) settings.
Our method thereby allows easy switching between a real pilot product and the virtual prototype, enabling, beyond many, application in use cases such as experimental design research, product usability testing as well as shopping or marketing tasks.
To do so, we treat the VOI detection task as an \emph{image classification problem} and thereby classify the individual fixation-relevant frames of the gaze replay videos using \emph{CNNs}  (cf.~Figure~{fig:cnn}).
In doing so, we simplify the three-dimensional task to a two-dimensional one.
Even though the neural networks is only trained with two-dimensional images, by training it with depictions of the same VOI from all perspectives, it can also recognize different perspectives on the same three-dimensional body.
To solve the problem of insufficiently available, scene specific, and annotated databases, we make use of the virtual model (cf.~Figure~\ref{fig:synda}).
Furthermore, we use a \emph{generative adversarial network (GAN)} as an image augmentation technique to adapt the training data to real environmental factors, and thus overcome the need for challenging photorealistic simulations.
In this paper we will give a proof of concept of our method, by carrying out semantic gaze analyses for eye-tracking-supported design reviews on a real coffee machine and its virtual prototype.
Among the data science based approaches, this is to our knowledge the first one that allows VOI annotation not only on an object level but also on a product feature level, while not relying on pre-annotated training data.
The main contributions of our paper are:

Contribution 1. We show how arbitrary new training data sets for the annotation of VOIs can be built up almost fully automatically doing image augmentations with Cycle-GAN and thereby represent the experimental world (virtual and even real environments) sufficiently accurate.

Contribution 2. We present a machine learning approach to annotate VOIs at a feature level and only on the basis of  synthetically generated training data, which reaches state-of-the-art accuracy, while allowing cross-platform use.

\section{Related Work}
\label{sec:related_work}

\subsection{Manual and Bounding Shape Approaches}

The naive way is to annotate data on the basis of scene videos using human annotators.
According to~\citeN{Holmqvist.2011} this can take up to 15 times the duration of the scene video.
Manual annotations of areas of interest (AOIs) and VOIs are therefore often supported by bounding shape approaches.
Most analysis software of eye-tracking hardware vendors provide such functions, to define the delimitations of the AOIs~\cite{Simko.2019}.
These bounding shapes are manually drawn, which is time consuming for dynamic stimuli, where they have to be defined frame by frame.
To reduce the manual bounding box definition efforts, \citeN{Kurzhals.2015} propose a method, which only requires to manually draw key bounding boxes in sparse video frames between key positions.
The bounding boxes of the remaining frames are interpolated linearly.
The detection of the AOIs is thereby supported by a cluster visualization.
Other approaches equip bounding shape approaches with feature tracking methods \mbox{\cite{Bertolino.2012, Bertolino.2014}}.
However, those approaches are just applicable for synchronized data within the same coordinate system, such as achieved from video based experiments with remote eye tracking.
When it comes to mobile eye-tracking data in 3D scenes
those approaches reach their limits.
This is because either the bounding boxes have to be defined individually for each video, or the fixation positions have to be manually transferred to 
annotated reference pictures.

\subsection{Tracker and Model-Driven Approaches}
\label{sec:tracker}

The basis of tracker and model-driven approaches are makers or motion-tracking systems, i.e. additional hardware.

A comparatively old, but widely applied approach is the use of
physical markers, such as infrared based markers.
Those are attached to plane surfaces within the experimental environment for automated mapping.
For an overview see~\citeN{Kohler.2011}.
Especially when it comes to AR applications, fiducial markers such as ARToolKit, ARTag, AprilTag or ArUco are commonly used, as the help to integrate synthetic content into the real world view at exact positions. This technique can also be applied to transform AOI visualizations from camera space to world space as shown in \mbox{\citeN{Duchowski.2020}}.
However, those approaches reach their limits, when it comes to interactive use cases involving 3D objects with non-planar and undercut surfaces. Here, many markers would be necessary to ensure their visibility from every possible viewing angles, while representing a disturbing factor as they expand the visually perceptible content (i.e. in product design evaluations tasks).
When it comes to interactive three-dimensional scenes, geometry-based approaches~\cite{Paletta.2013,Pfeiffer.2014,Pfeiffer.2016} define the state-of-the-art.
They rely on annotated geometrical models of the stimulus content.
The gaze data are then used to calculate the intersection of the VOIs in the proxy model and the ray of gaze.
To do so, the geometrical representation must be aligned with the real world, which is called registration.
For this an isomorphic coordinate system has to be generated using
markers or external tracking technologies.

\subsection{Computer Vision and Data Science Approaches}

In recent years methods from the fields of computer vision and data science became available.
Using those, it is increasingly possible to identify the visual content looked upon.
Thereby, images of the scene camera represent the basis of the semantic gaze analysis.
Automated methods are usually based on images of the scene video.
Some approaches
use feature extraction algorithms, such as Scale-Invariant Feature Transform (SIFT) or Speeded-Up Robust Features (SURF),
along with a classifier \cite{Toyama.2012, Brone.2011, Simko.2019}.
Others rely on CNNs~\cite{Sattar.2017, Steil.2018}.
For a detailed overview on neural networks in object recognition refer to~\citeN{Zhao.2019}.
However, a big unsolved problem of these approaches is that all of them require a huge database of scene specific images from various perspectives.
A manual setup of such a database, is very time consuming.
Yet, existing databases, such as~\cite{Deng.2009,Everingham.2010,Lin.2014} limit the granularity of classification to delimited objects (e.g. car, street, tree, dog) but do not offer to distinguish between their components.
As a solution~\citeN{Brone.2011} proposed a ``training-by-looking-at''-step to be done prior to the experiments.
Semi-automated approaches, such as~\cite{Pontillo.2010, Schoning.2016, Kurzhals.2017}, on the other hand, do not require any training data.
Here, the classification is carried out immediately on the experimental data. However, the classification proposed by the algorithm must then be presented to a human annotator, who interactively corrects proposed labels for active learning. 

 \begin{figure*}[t]
   \centering
   \includegraphics[width=\textwidth]{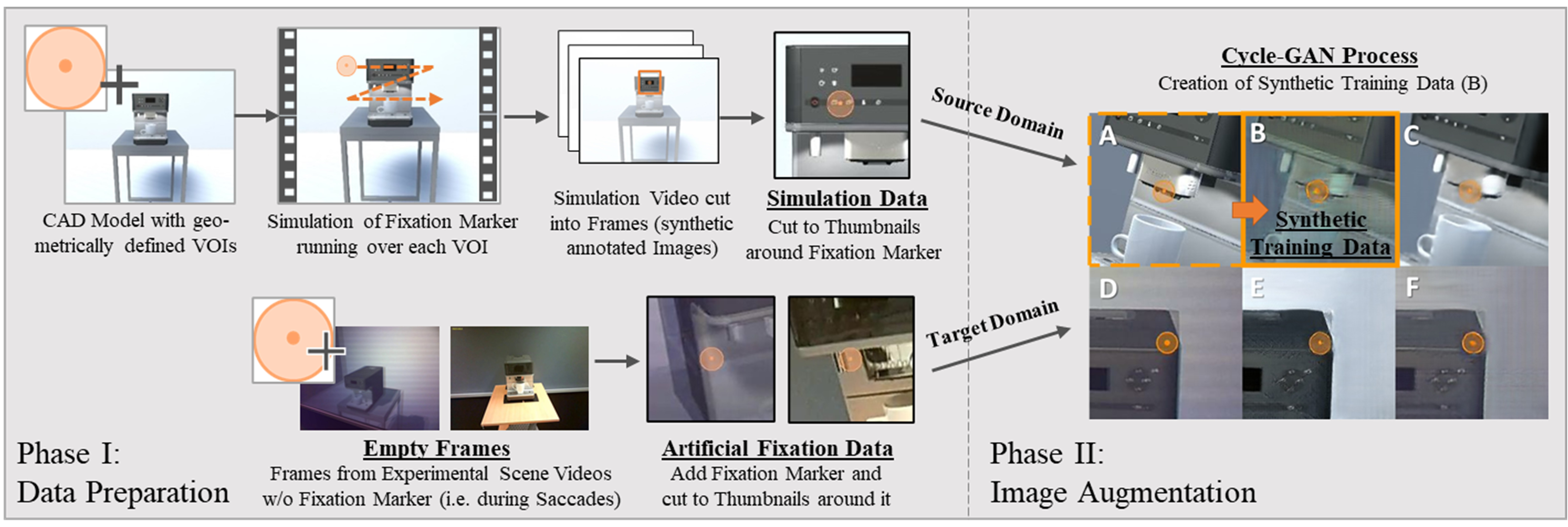}
   \caption{Creation of Synthetic Training Data using Cycle-GAN. In preparation of the network training simulation data and artificial fixation data are created (phase I). Those are then used as source and target domain to train Cycle-GAN models of the following architecture (phase II): (A) simulation image (source domain), (B) image A translated to target domain, (C) image B re-translated to source domain (cyclic A), (D) artificial fixation data image (target domain), (E) image D translated to source domain, (F) image E re-translated to target-domain (cyclic E). The transformation utilized for the channel-shift in our approach is the one from (A) to (B). It is used to create a synthetic training database for later classification network training.
   }
   \label{fig:synda}
   \Description[Creation of Synthetic Data]{long Test}
 \end{figure*}

\section{Own Approach}
\label{sec:own_approach}

Our approach tackles the
annotation problem described in Section~\ref{sec:introduction} using recent object recognition algorithms.
An overview of the proposed method is given in this section, as well as in the~Figure~\ref{fig:synda} and \ref{fig:cnn}. Methodological details are than given in Sections \ref{sec:prep} to \ref{sec:classification}. An application of the proposed method along with use case specific details on experiments and network trainings are presented in Section~\ref{sec:evaluation}.
The system is composed of the following main steps:

(1) Use a CAD model to prepare training data for the Cyle-GAN (Figure~\ref{fig:synda}).

(2) Use the Cycle-GAN to create a synthetic data set with real-world-alike appearance (Figure~\ref{fig:synda}).

(3) Use the synthetic data set to train a convolutional neural network (CNN) (Figure~\ref{fig:cnn}).

(4) Take the experimental data and predict the VOIs with the help of the trained CNN model. (Figure~\ref{fig:cnn}).

The essential resource for using object recognition algorithms is a suitable database.
To provide  an annotated database on a feature level, we make use of a CAD model or virtual prototype, in which we geometrically define the VOIs.
In a simulation a fixation marker is moved over all visible surfaces of the VOIs (cf.~Section~\ref{sec:synthetic}).
Henceforth, the frames exported from the simulation represent the annotated training database for the classification task (cf.~Figure~\ref{fig:synda}, left).

At the same time, we are capable to generate an approximately complete population of all possible perspectives, viewing angles and experimental situations, while getting rid of time-consuming manual effort to create suitable databases.
However, the creation of photorealistic data is  a challenging task~\cite{Mayer.2018}.
It is especially not obvious which aspects of the real world are relevant and must be modeled.
To overcome this obstacle, while avoiding intense simulation efforts, we propose
an image augmentation technique called Cycle-GAN to achieve photo-realistic optics
in an highly automated process (cf.~Section~\ref{sec:channel}).
Thereby, we perform a channel shift to transform the simulation data into images that appear to look like the pictures of the desired target domain (cf.~Figure~\ref{fig:synda}, right).
On those synthetic data, we train a classifier (cf.~Section~\ref{sec:classification}).
The resulting CNN model is finally used to predict the fixated features during real experiments and to annotate the VOIs (cf.~Figure~\ref{fig:cnn}).

\subsection{Preparing Data for Classification}
\label{sec:prep}
For our approach we rely on two different data sets.
The neural networks are trained on synthetic data described in the next section.
In this section we show how the experimental data, which will later be classified upon, are preprocessed.

\subsubsection{Experimental Data}
\label{sec:hauptstudie}
To prepare our experimental data for the image classification task, we rely on egocentric videos, which are split into frames.
Thus, our approach requires scene videos, which is available for all eye trackers that include a field-of-view camera.
These scene videos are usually enriched with gaze events (fixations and saccades).
The resulting visualization is called gaze replay~\cite{Holmqvist.2011}.
In contrast to scene videos prepared for manual annotations, for our approach the fixation markers should only be displayed over the actual fixation duration and without connection paths (i.e. saccade representation).
It is thus guaranteed, that only one fixation marker is contained in each frame.
However, a data export with gaze coordinates mapped to the coordinate system of the scene video and gaze event classification can be used alternatively to the gaze replay.
Subsequently, the gaze replays are divided into individual frames.
Thereby, only frames that fall within the time span of a fixation have to be considered for the further annotation process.
To perform this task, we rely on a data export, from which the times of the gaze events, or respectively at minimum the start and end time of the fixations, can be derived.
As a result, the videos recorded in the experiments have now been decomposed into classifiable images (cf.~Figure~\ref{fig:cnn}).
We refer to these as \emph{experimental data}.

\subsubsection{Computational Feasibility}
\label{sec:preprocessing}

As the data is derived from gaze videos, we face high resolution images.
Since the image size grows quadratically, it is neither temporally nor economically or ecologically reasonable to work on full size images.
Thus, the frame size has to be vastly reduced.
For this, images can be resized or cropped to relevant details.
While resizing provides additional scene information, it entails the obstacle that the relevant area, which is tagged by the fixation marker is represented with less pixels.
Thus, by resizing relevant information about the fixation marker and its highlighting area are lost.
Cropping on the other hand might ensures that information encoded in the pixels of the relevant area remain unchanged.
Additionally, in the case of cropping to relevant image sections, additional preprocessing is provided as the fixation marker is positioned in the center of the image and bloating information about background features are eliminated.
In consideration of our task (cf.~\ref{sec:evaluation}) and in accordance with a pretest, we
decided to crop to an image section in the vicinity of the fixation, as this section represents the currently watched region~\cite{Holmqvist.2011}.
The resolution is thereby determined by the eye tracker's scene camera.
One should always opt for the best resolution here, as this helps to determine the watched VOI, when fixations are close to VOI borders.
As the cropping strategy is commonly used in research~\cite{Kurzhals.2017, Steil.2018},
we adopt the pre-existing term of \emph{thumbnails}.
To decide on the thumbnail size, it has to be ensured that relevant features around the viewed region are covered.
For this, the average viewing distance has to be taken into account, i.e.\ the further subject and object are distanced, the more features are depicted in the recorded images and thus it can be cropped to smaller thumbnails.
We thus choose the smallest possible thumbnails, such that in the majority of cases at least one neighboring VOIs is visible.
Modern eye-tracking systems provide $x$- and $y$-coordinates of the fixation marker, i.e. the exact position of the foveal fixation.
Those can be used for cropping the images to thumbnails, such that the fixation marker is in the center of the image.

\subsection{Generation of Synthetic Training Data}
\label{sec:database}

In general, CNN-based classification of data 
 is more promising, the more representative the training data set is for the experimental data.
Thus, the next sections deal with the creation of data sets to train the classifier upon.
A challenge, that our approach faces, is that we use a purely synthetic and automatically created data set to train the model, while later classifying real experimental data.
Thus, the synthetic features, which the fixation marker is placed upon in the training stage, should represent the real-world features as closely as possible.
This is achieved through a combination of simulation (cf.~Section~\ref{sec:synthetic}) and image augmentation (cf.~Section~\ref{sec:channel})

\subsubsection{Simulated Data}
\label{sec:synthetic}

In order to train our classification network exclusively on synthetic data, we rely on a virtual model, which resembles the presented stimulus reasonably.
In our approach a CAD model, enriched with textures, as well as light sources and shadow simulations is built.
The VOIs are defined in the virtual model using simple cubic or cylindrical shapes suiting the volumes.
Next, a marker is created, which resembles the fixation marker of the gaze replays.
This marker is then systematically moved across the surfaces of each VOI (cf.~Figure~\ref{fig:synda}).
This is done by defining a moving path for the marker and exporting the resulting simulation. As the VOIs resemble volumetric bodies, which can be seen from different perspectives, usually several surfaces of the VOI have to be taken into account, certainly those, to which the view is not constructively obstructed. If the experimental scene is designed to be interactive, surfaces that can become visible through interactions must be taken into account, too.
Additionally, the same process has to be done with non-VOI surfaces such as model surroundings.
Those images will later present the default class if no VOI is looked at.
The resulting simulation is exported as a set of frames to build the synthetic database.

As we are dealing with interactive three-dimensional stimuli and freely moving participants, VOIs can be seen from various perspectives.
Thus, a sufficiently large number of different viewing angles has to be created.
To accommodate different viewing directions, the virtual space camera is placed in variable positions relative to the vertical axis of the model.
Which positions have to be taken into account is strongly dependent on the movement space available to the subjects during the trials \mbox{(cf.~Section~\ref{sec:study})}, i.e. if subjects move around an object, it is beneficial to use uniform angular distances.
Moreover, tilting of the head can be included by rotating the simulated camera.
The distance of the camera should thereby be chosen equal to the average distance of the participant to the target objects and features recorded in the experimental data.
Furthermore, the heights of VOIs with exceptionally high attention-attracting capacity should be generated.
This viewing direction is considered additionally, knowing that people approach important features in order to get a more direct view.
For simulation detail on our use case refer to \mbox{Section~\ref{sec:simulation}}.
In addition, the virtual space camera is to be placed at mutable eye heights to take different body sizes into account.
This is done with respect to the standardized human measurements defined in \mbox{DIN 33402-2}~\cite{DeutschesInstitutfurNormung.200512}, by taking the eye level heights of the 50-percentile and the 95-percentile for men (which correspondents closely with the 95 percentile for women) as well as the 50-percentile for women (which correspondents closely with the 5 percentile for men) in the age span from 18-65 years into account for the simulation.
Thus, a decent representation of viewing angles from the experimental data is achieved.
However, the reader is advised to check the degree of compliance of resulting viewing angles with those of the experimental data to be analyzed.
It must be considered, that the given recommendations always depend on the individual characteristics, such as the heights of the participants, and may not consider particularities of any given experimental setup.
As this process can be conducted primarily automated the manual effort is reduced to the initial definition of VOI and possible camera positions.
In the following we refer to these generated images as \emph{simulation data}.

\subsubsection{Image Augmentation}
\label{sec:channel}

The goal of our approach is to be able to carry out VOI classifications without investing considerable effort in the generation of a manually annotated, experiment-specific database.
Therefore, we decided to generate our training database using a simulation of the experimental scene.
As noted in~\cite{Mayer.2018}, it is difficult and additionally enormously time-consuming to produce realistic simulations.
In accordance therewith, instead of investing a lot of time in simulations, the virtual model is only enriched with simple elements such as textures, lighting and shadows as described in Section~\ref{sec:synthetic}.
Subsequently, a domain transfer is carried out using generative adversarial networks (GAN) in order to adapt the simulation images to the appearance of the experimental data .
This has the advantage to work fully automated and thus reduces the simulation effort as it does not have to depict every aspect of reality.
For this, our approach uses a method called \emph{Cycle-GAN} from~\citeN{Zhu.2017}.
This method can be used to transfer images between two different domains, in the absence of
paired examples.
In our case Cycle-GAN translates an image from the domain of the simulation data (\emph{source domain}), to the domain of the experimental data (\emph{target domain}).
The data basis thus achieved is referred to as \emph{synthetic data} (cf.~Figure~\ref{fig:synda}).

\begin{figure*}[t]
  \centering
  \includegraphics[width=\textwidth]{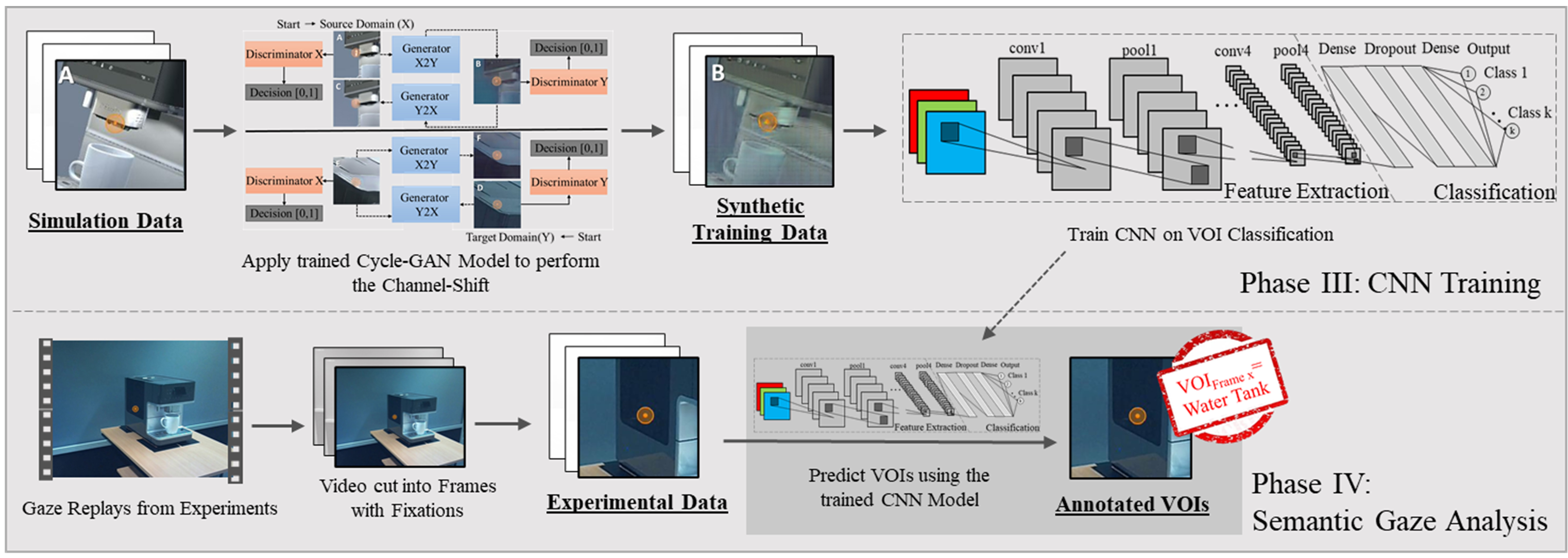}
  \caption{Visualization of the classification model training (phase III) and the flow of data up to the semantic gaze analysis (phase IV). }
  \label{fig:cnn}
\end{figure*}

\subsection{Image Classification}
\label{sec:classification}

With the synthetic database at hand, we can now address the annotation problem described in Section~\ref{sec:introduction}.
Most object detection approaches broadly consist of an object localization combined with an image classification, which allocate pixels to instances by means of adjacent pixels that share textures, colors, or intensities.
Both tasks are particularly difficult on a feature level, as the boundaries of the individual features are often not distinguishable by hard contrasting edges.
However, we can make use of the advantage, that the eye tracking data already provide us with the exact coordinates of the fixation relative to the gaze replay and hence we know exactly, which area of the image has to be considered for the classification.
The recently popular but computationally more expensive methods of semantic or instance segmentation can therefore be dispensed with, as the localization of each VOI in the image is of marginal importance while increasing computational costs.
By also requiring additional simulation effort, those methods become both, uneconomically and time consuming.
We thus tackle the problem as an object recognition task.

The image classification is done using the neural network architecture ResNet50v2 from~\citeN{He.2015}.
We decided to choose this architecture based on empirical experience
and as it is a state-of-the-art approach in image classification, as current papers such as~\citeN{Touvron.2019} show.
The general network architecture is modified such that the input layer corresponds to the size of the thumbnails.
Additionally, the final layer is replaced by a dense layer consisting of as many neurons as the number of VOIs to be classified. This layer is equipped with a softmax activation function.
The training data is generated as described in Section~\ref{sec:synthetic}.
Emerging from the simulated viewing points and differently sized VOIs, an over-representation of prominent VOIs in the synthetic data set is expected, resulting in unbalanced sets of training data.
To compensate for this, the smaller classes should be oversampled in the training stage.
To enrich the variety of the training data set, image augmentations are applied, such as rotations, translations, and color shifts.

\section{Evaluation}
\label{sec:evaluation}

To evaluate the approach presented above, we check upon our main contribution goals (cf.~Section~\ref{sec:introduction}), whether
\begin{itemize}
\item[C1.] an almost fully automatically created synthetic database using Cycle-GAN can represent the experimental world (virtual and even real environments) sufficiently accurate.
\item[C2.] a machine learning approach to annotate VOIs at a feature level and only on the basis of synthetic training data reaches results that can compete with state-of-the-art approaches both in real and virtual application cases.
\end{itemize}
To do so, our approach has been applied and evaluated based on an experimental setup in the field of real and virtual prototype testing.
Thereby, the application of eye tracking is aimed to track the focus of interest of a person reviewing a product.
For more information on the original experiment of the study,
refer to~\citeN{Blackert.2019}.
In this work, we will only discuss the experimental information relevant for the application of our approach.

\subsection{Experimental Setup and Baseline}

The proposed semantic gaze analysis is applied to \emph{two data sets}, both derived from the user study described in the following subsections.
Thereby, a fully automated coffee machine serves as the experimental object.
While the first data set contains eye-tracking data acquired from inspecting  the physical coffee machine (\emph{real-world setting}), the second data set is derived from gaze data on a stereoscopic three-dimensional virtual prototype (\emph{virtual-reality setting}).
The VOI definition is carried out manually and in accordance with the geometries of the product features defined in the CAD model (cf.~Figure~\ref{fig:VOI}).
We perform the analyses for both settings, to show the range of application of the proposed semantic gaze analysis, ranging from experiments in virtual environments over arbitrary XR setups to laboratory real world experiments.
It further enables us to estimate how much the annotation result is influenced by the degree of similarity between the virtual model and the stimulus under experimental conditions.
As we consider the virtual prototype, which is derived directly from the CAD model, to be much more similar to the simulation data, it is expected to show a better overall performance.

\subsubsection{Conditions / Baseline}
\label{sec:condition}

We investigate the performance of our proposed semantic gaze analysis approach by comparing our algorithm to \emph{EyeSee3D}, an approach proposed in~\cite{Pfeiffer.2014, Pfeiffer.2016}.
Similar to our approach, it requires a three-dimensional geometric model. We thereby compare approaches both offering a solution for the same field of applications.
To receive \emph{ground truth} results for a fair method comparison, we rely on the results of \emph{manual annotation}.
With the ground truth being set, we calculate the weighted \emph{precision} and \emph{recall} as well as the \emph{weighted F1-Score} for each method. The \emph{accuracy} corresponds to the recall in the weighted case.

\subsubsection{User Study Design}
\label{sec:study}
A total of 24 participants (6 female and 18 male) between the age of 23 and 62 (M = 30.65, SD = 9.13) took part in the experiments. First, participants are introduced to the experiments and asked for voluntary participation and anonymized data use approval. Afterwards, a 3-point calibration of the eye-tracking system is performed with each participant.
Next, all subjects are asked to interact with the product in both the virtual and the real settings, however in counterbalanced order.
The experiments are divided into two phases.
In the first phase, perception is studied by asking the subjects to freely explore the object for 60 seconds. This includes free movements around the machine, whereby the average distance to the object is set at around one meter for both settings.
However, the semicircle, in which one is allowed to move around the machine, is limited to $\pm 45 ^\circ$.
This prevents distortion effects in the virtual representation otherwise resulting from attempts to look behind the machine.
In the second phase of the experiments the subjects are asked about their perceptual impressions.
Thereby, their attention is led to certain product features as they have to solve tasks such as brewing coffee.
Due to the free movements of the test persons around the front and the sides of the product, we deal with subject individual scan paths and scenes. This combination is particularly challenging to analyze.

\subsubsection{Apparatus}
We use Unity3D to prepare the virtual prototype on the basis of a CAD model and present it ste\-reo\-sco\-pi\-cal\-ly three-dimensional on a 100-inch Powerwall using two projectors with a resolution of $1920 \times 1200$ pixels each.
In parallel, we record the subjects eye movements with mobile eye-tracking glasses of Senso Motoric Instruments (SMI) at a 60Hz binocular sampling rate.
The scene cameras video resolution is set to $1280 \times 960$ pixels, the corresponding frame rate is 24fps.
To ensure a stereoscopic three-dimensional vision in the virtual setup, we equip the glasses with polarizing filters.
Additionally, the glasses are equipped with SMIs 3D-6D head and eye-tracking system including the optical outside-in motion tracking system OptiTrack Prime$ ^{\text{x}} $13W and the associated motion capture software OptiTrack Motive 1.10.2.
This allows the recording of the users' head position and orientation within a defined space.
To access motion capturing and eye-tracking data we apply the SMI iView VRPN Server included in SMIs iView ETG 3D-6D SDK 2.7 and SMI iView ETG Software 2.7.
Furthermore, we use BeGaze 3.7 to classify relevant gaze events (saccades, fixations, and blinks) and to generate video-based gaze replays.

 \subsubsection{Evaluation of the Tracking}
 \label{sec:tracking quality}

\begin{figure}[t]
  \centering
  \null
  \hfill
  \begin{minipage}{.5\textwidth}
  \includegraphics[width=\textwidth]{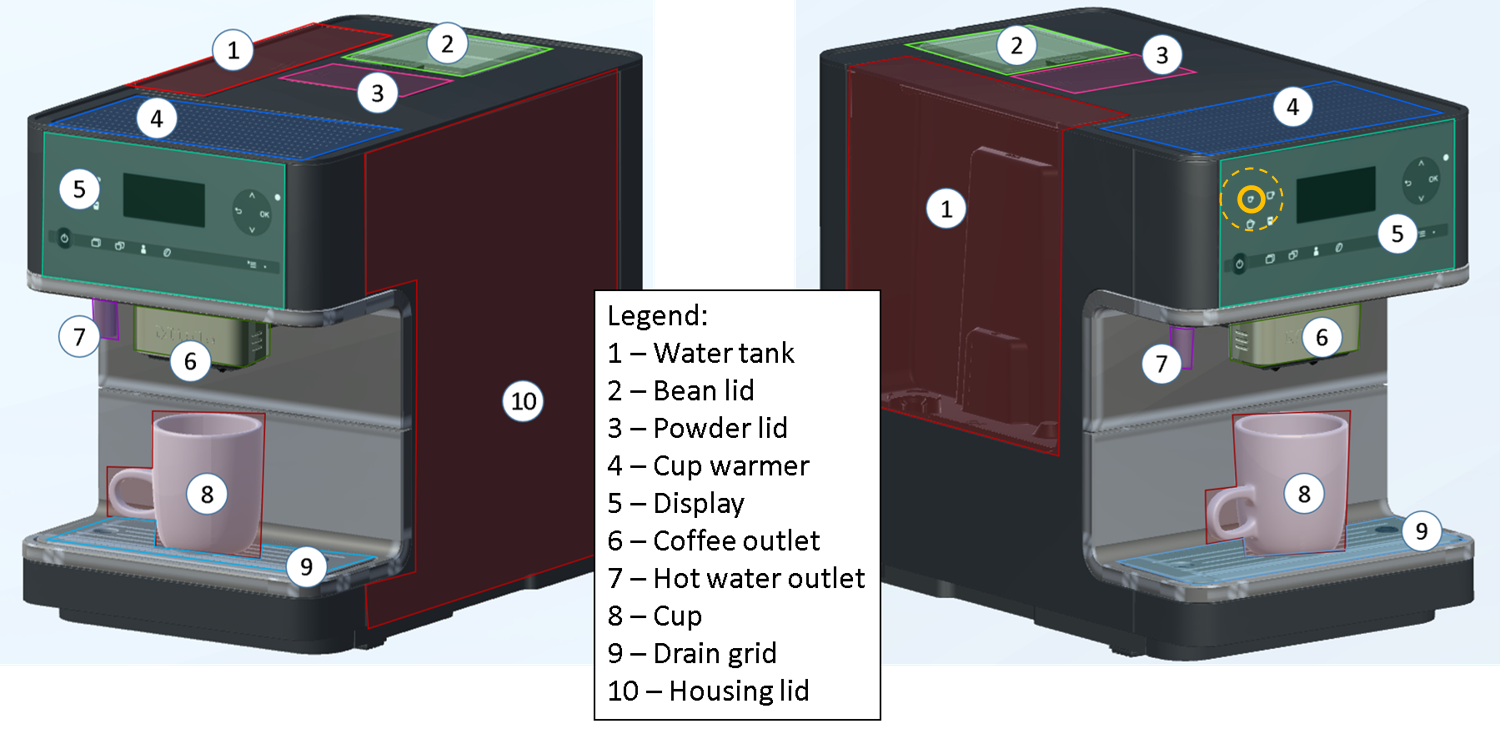}
  \caption{VOIs as defined in the case study.}
  \label{fig:VOI}
  \end{minipage}
 \hfill
 \begin{minipage}{.2\textwidth}
 \includegraphics[width=\textwidth]{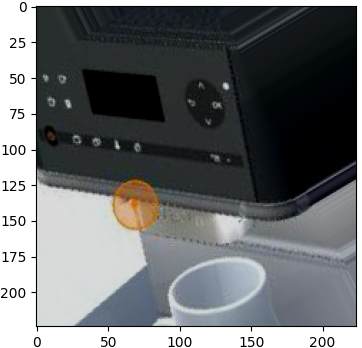}
  \caption{An obstacles for the semantic gaze analyses are unambiguous VOIs.}
  \label{fig:obstacles}
 \end{minipage}
 \hfill
 \null
\end{figure}

 When it comes to eye-tracking data quality, there are mainly two critical aspects to judge upon.
 The first one is \emph{accuracy}, which describes the average difference between the real stimulus position and the measured gaze position.
 The second one is \emph{precision}, which gives an estimation about the ability of the eye tracker to reliably reproduce the same gaze point measurement.
 To give an estimation about those measures we rely on a button on the coffee machine that participants have to revisit several times throughout the experiment.
 Thus the button serves as gaze target area (cf. yellow marking in Figure~\ref{fig:VOI}).
 With a diameter of 10mm, this area corresponds to a $0.6 ^\circ$ viewing angle, which complies approximately with the accuracy specification of the tracking given by the manufacturer.
 In order to only consider reliable tracking data, a $1.5 ^\circ$ target area is defined around that button, in which all forced revisiting fixations have to occur.
 This condition is met by 42 of the 48 data sets. For the evaluation of our approach, we will thus only continue to consider these data sets.
 The inaccuracies of the remaining data sets could be attributed to a slight slippage of the glasses, e.g. by touching them during the experiments, and can be compensated by an offset correction.
 This would allow a subsequent annotation with our proposed method, but not for the live annotation using EyeSee3D.
 The data exclusion is therefore retained to ensure fair comparability of the methods.
 In respect thereof, we also check the motion tracking coverage rate, as the geometric-based EyeSee3D approach is strongly dependent on stable motion tracking.
 For that, we examine, if poses are identified during each time span in which a fixation is detected.
 This is the case for 99.65\% of all fixations, which the authors perceive as sufficiently high.

\subsection{Data Preparation and Network Training}
\label{sec:preliminary}

Building upon the data basis discussed in the last section, we now prepare our training data for the classification task and  train the neural networks using these data.

\subsubsection{Generation of Simulation Data}
\label{sec:simulation}

As described in Section~\ref{sec:synthetic} our simulation data are produced in
reference to the standardized human measurements. In our use case we have the
display as VOI with exceptional high attention-attracting capacity.
Therefore, we further include the display height as well as 30 cm above and below in order to mimic stooping for a better view on the machine.
The generated images are in a resolution of $1280 \times 960$, as this matches the resolution of the recorded SMI scene videos.

\subsubsection{Details of the Network Trainings}
\label{sec:training}

When it comes to network training, the first step is cropping the images to thumbnails.
As we deal with a viewing distance of around one meter, we decided to go with a thumbnail size of $224 \times 224$ pixel.
This guarantees that the shapes of the different VOIs as well as their surroundings are still recognizable.
The data size is thus reduced to $4.08\%$ of the original image size, yielding in computational feasibility, as discussed in Section~\ref{sec:preprocessing}.
Subsequently, the two Cycle-GAN networks are trained as described in Section~\ref{sec:channel}.
Both Cycle-GANs use 1000 images from the simulation data as source domain.
The first network additionally uses 1000 images recorded in the virtual-reality experiment as target domain, cropped to thumbnails.
The second Cycle-GAN instead uses 1000 thumbnails gathered in the real-world setting.
Refer to the image augmentation section (right) of Figure~\ref{fig:synda} for a demonstration  of such a domain transfer where an image in the source domain is depicted as \textbf{A} and the target domain as \textbf{B}.
We use a preprovided implementation\footnote{\url{https://github.com/LynnHo/CycleGAN-Tensorflow-2}} for the Cycle-GAN.
Except changing the numbers of epochs to 50, we do not branch from the standard parameters as described in the original paper~\cite{Zhu.2017}.

Using the so trained models, we can now generate our training data for the image classification.
All images are in thumbnail form.
We rely on a total of 100,000 simulated training images.
Those are augmented using both trained Cycle-GAN models.
Thus, we create 100,000 additional images in each of the two domains (real world and virtual reality).
A data set of 300,000 images (simulated data, real-world data and virtual-reality data) is consequently generated for the upcoming training of the classifiers, compare to Figure~\ref{fig:cnn}.
Each image is thus contained three times in the training set, however differing in its domains.
By doing so, we achieve a higher generalization of the classifying neural network.

The classification is carried out as described in Section~\ref{sec:classification}.
As training data for the network we use the data discussed in the last paragraph.
We train the ResNet50v2 network exactly as described in the original paper~\cite{He.2015}.
As we use thumbnails of size $224\times 224$ we do not have to modify the input layer. The output layer is replaced by a layer consisting of as many neurons, as VOIs are to be classified.
In our cases we deal with 12 classes; the 10 VOI classes depicted in Figure~\ref{fig:VOI} as well as two default classes (``coffee machine but no VOI'' and ``no coffee machine'').
We train the network using the Adam optimizer and a learning rate of 0.001 over 20 epochs with the sparse categorical crossentropy as loss.
Furthermore, we use popular vote to annotate the fixations as, due to sampling rates and fixation duration, fixations consist of scanning points (EyeSee3D) or frames (CNNs).

\subsubsection{Computational Costs}
\label{sec:computational-costs}

All our computations are carried out using a GPU with an Nvidia GeForce RTX 2060 SUPER chip and 8 GB GDDR6-RAM.
The training of  the Cycle-GAN networks takes around 7 hours each while training the classification ResNet50v2 takes a total of 9 hours.

\subsection{Results and Discussion}
\label{sec:discussion}

As the experiment is conducted on two different domains, i.e. in a real-world setting and a virtual-reality setting, our evaluation is divided into two parts, compare to Table~\ref{tab:results}.
As we deal with an unbalanced multi class classification problem, we use the weighted average precision and recall (which is equivalent to the accuracy) as well as the weighted F1-score.
Thereby, the over-representation of the large classes in the test data set is compensated, i.e. only predicting the largest class is not rewarded.
This is a common technique for unbalanced multi class problems.

\begin{table*}[t]
  \centering
  \caption{Comparison of our approach (4) to two models (1, 2) showing the performance share of the image augmentation using Cycle-GAN and the synthetic data set, as well as a comparison to our baseline EyeSee3D (3).}
  \label{tab:results}

  \centering

  \begin{tabular}{l|ccc|ccc}
    \toprule
    &\multicolumn{3}{c|}{\textbf{Real world}}&\multicolumn{3}{c}{\textbf{Virtual reality}}\\
     & Precision &  Recall/ & F1-Score &  Precision &  Recall/ & F1-Score\\
    &  & Accuracy  & &  & Accuracy  & \\
    \midrule
    1) ResNet50v2 (human annotations) & 0.33 & 0.34 & 0.33 & 0.38 & 0.37 & 0.35\\
    2) ResNet50v2 (synthetic) & 0.32 & 0.19 & 0.14 & 0.41 & 0.32 & 0.31 \\
    \midrule
    3) EyeSee3D  & \textbf{0.60} & 0.58   & 0.56 & \textbf{0.66} & 0.59 & 0.56 \\
    4) ResNet50v2 (synthetic + Cycle-GAN)      & 0.59 & \textbf{0.59}  & \textbf{0.58} & 0.63 & \textbf{0.61} & \textbf{0.61} \\
    \bottomrule
  \end{tabular}

\end{table*}

\subsubsection{Results of the Semantic Gaze Analysis}
\label{sec:results_SGA}

Generally, it can be said that both approaches, our neural network framework and the geometrical EyeSee3D approach, yield comparable results.
In both settings our approach reaches better results in recall/accuracy then EyeSee3D, while EyeSee3D yields slightly better results for the precision.
Still, our approach achieves a higher F1-score, which is used in order to balance recall and precision.
Overall, the CNN-approach performs slightly better in virtual reality then in the real world.
This is the expected behavior as visual aspects of the virtual reality are closer represented by the simulation data, while in the real-world scenario environmental influences such as reflections complicate the task.
The authors however expect both scenarios to increase the performance, if the image augmentation using Cycle-GAN can be further improved.
We draw this conclusion as we saw geometrical and visual modifications performed by the Cycle-GANs concerning relevant features.
However, as results in Section~\ref{sec:results_synthetic} demonstrate, Cycle-GAN already vastly compensates the \emph{reality deficit}, which synthetic data generally suffer from.

\subsubsection{Reasoning on the Synthetic Data Approach}
\label{sec:results_synthetic}

To evaluate the performance share of the image augmentations, we train a ResNet50v2 architecture on the unaugmented simulation data set exclusively, i.e. on the underlying 100,000 simulation images.
The results are listed in Table~\ref{tab:results}.
As expected, the classification performance of the so trained neural network is significantly weaker than the performance of the network trained on the synthetic data with the Cycle-GAN image shift.
The unaugmented data set does not seem to represent the experimental data sufficiently, especially in the real world.
Contrarily, the proposed image augmentation using Cycle-GAN seems to be able to compensate these reality deficits to a sufficient degree.
We come to this conclusion, as the network trained on the augmented data outperforms the instance without the augmented data by almost doubling its accuracy in the virtual-reality setting while improving the results in the real world by an even larger degree.

An alternative way to overcome the reality deficit would be to generate a scene specific data basis from scratch using hand-annotated images.
As discussed before though, generating broad and high quality databases for training CNNs manually is a time intense and economically highly inefficient task.
To demonstrate this, we do a comparison to the \emph{human annotator method}, which is still the most-used way to generate training databases for neural network approaches.
Thus, we generate a database of 30,000 manually annotated images from scene videos.
This process takes around 25 hours if the annotation rate is as fast as 20 images per minute.
Using the same hyper parameters as in our ResNet50v2 classification, we train a second instance of this neural network architecture on these manually annotated training data.
Even though the generation of the training data takes around 10 hours longer than in our approach and requires a drastically higher human effort, the so trained model only achieves an accuracy of 34\% in the real world and an accuracy of 37\% in the virtual reality, compare to Table~\ref{tab:results}.
It can thus be seen, that our network trained on the synthetic data achieves a much higher accuracy at lower time input, which also makes it economically more viable.
Therefore our empirical experiment seems to justify the assumption, that it is not efficient to manually annotate a data basis from scratch.
While it is very time consuming, the database is not broad and precise enough to train a complex neural network such as ResNet50v2.
Moreover, the network is likely to overfit on small data sets.
Ideally, one would repeat the same experiment with a manually annotated dataset of the same size as the computer generated one.
In this case, the authors expect the neural network trained on the manual annotations to outperform our approach.
Because of the high manual effort for conducting user experiments and annotating data this is however hardly feasible.

\subsubsection{Discussion}
Nevertheless, the results of our approach, might initially look weaker than expected from a practitioners' point of view.
However, when judging the results certain aspects have to be taken into consideration.
First of all, we rely on manually annotated data as a \emph{ground truth}.
As human annotators are prone to biased annotations, a perfect annotator agreement can never be hoped for.
Experiments on the annotation agreement such as in~\cite{Pfeiffer.2016} give reason to expect an agreement of around 80\% between two human annotators as an expectable optimum.
To better understand this problem, we advise the reader to look at Figure~\ref{fig:obstacles}.
There the fixation marker is ambiguously located between four different VOIs and default classes some of which are adjacent and others which are simultaneously hidden due to depth effects.
Thus, if the ground truth is annotated by a human, as in our case, one should never expect to achieve an optimum of 100\%.
Furthermore, both approaches deal with a problem which is caused by the SMI BeGaze fixation detection algorithm, used during data aggregation.
This phenomenon is called ``running fixation'' and is also described in~\cite{Pfeiffer.2016}. Thereby, the gaze cursor is not stable on a single VOI during the whole fixation duration but moves over different VOIs.
Overall, it can thus be said that we can not act on the assumption of a perfect ground truth.
Even a perfect method might therefore not be able to achieve an accuracy far beyond 80\%.

\begin{figure}
\centering
    \begin{tabular}{m{20em}m{20em}}
    \includegraphics[width=.3\linewidth]{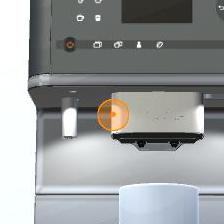}
    \includegraphics[width=.3\linewidth]{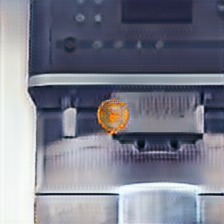}
    \includegraphics[width=.3\linewidth]{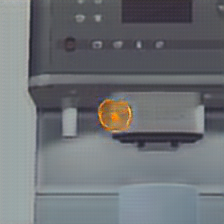} &
    \includegraphics[width=.3\linewidth]{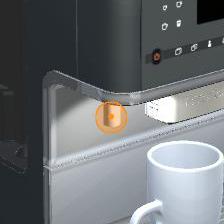}
    \includegraphics[width=.3\linewidth]{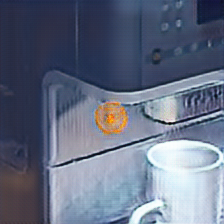}
    \includegraphics[width=.3\linewidth]{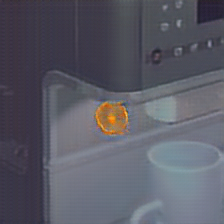}
    \end{tabular}
    \caption{Two additional examples demonstrating the domain transfer. The left image depicts the simulation while the middle and the right image depict the domain transfer into the real-world and the VR-environment respectively.}
    \label{fig:limitations}
\end{figure}

However, this fact alone might not explain the results.
As a closer look at the metrics for each VOI class reveals, our approach still suffers from some challenges.
While some VOIs such as the display are classified to high satisfaction, others such as the powder lid, show a weaker stability in classification.
For the display this might be traced back to high contrast edges between it and its surrounding housing.
The power lid however does not have such contrasting edges and is thus harder to distinguish from its surrounding.
The classification of such non-standalone objects is however not only a problem of our approach, but a general problem of classification using neural networks, as those are known to work better on detached objects then on a detailed feature level.
In general, the wrong assignments did not follow from random predictions but from predicting the default classes.
Moreover, it is surprising to see, is that even though in virtual reality and real world the powder lid is not detected well, in the real world it is still more often assigned to one of the default classes, while in virtual reality it is often mistaken for the bean lid.
This indicates, that the classification is further complicated as some shapes are only differing slightly, such as coffee powder lid and bean lid.
While this is expectable in product design as usually uniform design languages are followed, in other VR applications this challenge might not accrue.
Furthermore, VOIs might be covered behind other parts and only be partially visible while consisting of transparent or reflecting materials.
Our approach tackles this by using the simulation together with the image augmentation technique.
However, Cycle-GAN can also impair the quality of the images.
An example for this phenomena can be seen in the right example of \mbox{Figure~\ref{fig:limitations}}.
The marker is on the hot water outlet that is clearly visible in the simulated image on the left.
However, in the images after the domain transfer, i.e., the middle and the right image, the hot water outlet is barely visible because of artifacts generated by the Cycle-GAN.
Still, the marker is needed for the neural network to determine the position of gaze.
A possible approach to solve this is to omit positioning the marker directly on the image and provide the network with additional input variables for the $x$- and $y$-coordinate.
However this might not necessarily result in a better classification because of the large number of input variables for the image.
Results nevertheless show that our work can already compete with a state-of-the-art approach in two different domains.
Furthermore the comparison reveals that, due to the different operating principals of the systems, they also struggle in different ways. While EyeSee3D mistakes VOIs with other VOIs close by, our approach primarily mistakes VOIs with similar appearances.
Still, we are convinced that the scores of the approach have potential for being improved.
Additionally, carefully combining both approaches might further boost the performance.

Overall, the approach has high potential to be applied in agile product development and experimental prototype testing, as those fields typically rely on virtual product models.
For example it enables all of early prototype testing to be done in virtual-reality settings while allowing switching between a real pilot product and the virtual prototype in later stages.
Thereby, the tracking devices can even be interchanged in the course of the experiments as our approach only relies on gaze replays which are provided by all major eye-tracking vendors.
This makes it possible to switch between devices such as mobile eye tracking used in combination with Powerwalls, CAVEs or real prototypes and HMD integrated eye tracking.
Furthermore, the described fields also deal with incremental changes in products that might have to be evaluated in iterative steps.
Caused by the nature of iterative virtual product development which deals with minor modifications of product designs between subsequent iterations, the training of the neural networks does not have to be repeated from scratch for each new iteration.
Instead the existing models from previous iterations can be fine-tuned for the new product design changes.
Some layers of the neural networks might be frozen in this context, which resembles a recent technique known as \emph{transfer learning}.
Such methods help to lower the computational costs that come with the training of such deep and complex neural networks.
It can also be reported that complex, extensively trained, neural networks may not always deliver the best results.
This is especially true for Cycle-GAN, where we achieve better results using 50 epochs instead of 200.
The authors suspect that this is caused by overfitting effects that occur from homogeneous training data.
In the case of using complex architectures such as ResNet50v2 or Cycle-GAN, we advise the reader to train the models on few epochs first to evaluate whether a training on more epochs is sensible.

\subsubsection{Limitations}
The proposed method can easily be applied to pre-modeled environments.
However, in use cases, in which a digital twin would have to be built first (e.g. field experiments), the method is limited.
In these cases the efforts involved in building up the digital representation might not justify leaving the previously known paths to annotate VOIs \mbox{(Section~\ref{sec:related_work})}.
Another limitation is, that while the study gave a proof of concept for two different domains, it is not yet a full demonstration of the generalizability of the approach.
The study should thus ideally be repeated with additional use cases beyond the coffee machine.
However, this requires that a new user study has to be conducted and a database with annotated ground truth labeling has to be collected, which is a task with high manual effort.
The limitation likewise applies to the achieved precision and recall, which are not impressively high yet.
Therefore, the proposed method might not be reliable enough yet to be used in daily research projects without double-ckecking the results.
Finally, like all approaches with rigid classes, our approach shares the problem of classifying ambiguous cases, i.e., cases where the marker is in-between classes.
This problem can not be fully eliminated, because the annotated ground truth struggles from the same problem.

\section{Conclusion and Future Work}
\label{sec:conclusion}

In this work we propose a method for semantic gaze analysis using machine learning, while eliminating the resource-intense process of human annotations.
To do so, we rely on a reasonable big and scene specific data basis, which we generate using a basic simulation together with an image augmentation technique called Cycle-GAN.
Thus, we exclusively rely on synthetic data.
In doing so, our method especially suits applications for which a virtual model is already accessible.
This enables the approach to be used in all virtual and augmented reality use cases, as well as in experiments with real products for which a virtual representation exists.
We demonstrate the feasibility of our method by applying it to annotate data derived from an experimental design study. Those are conducted in two domains, one being a real-world and the other being a virtual-reality setup.
Thereby, we show that our approach can not only be broadly applied but can also already compete against state-of-the-art methods in semantic gaze analysis, while still having potential for future improvements.

A particular strength of our approach is, that in comparison to other methods for semantic gaze analysis, neither markers nor motion tracking systems are required.
This minimizes the risk of errors through latencies and registration failures.
Against manual or semi-automated annotation approaches, our method furthermore generates reproducible results while it does not contain a personal bias and is thus not prone to evaluator effects.
Our approach is particularly suited for fields of applications that can benefit from shifting between virtual, augmented and real-world settings.
For example in agile product development the built models, remain flexible to include iterated product developments and design changes from the first CAD model through virtual prototypes up to the real product.
Meanwhile, the same methodical evaluation can be used across platforms, i.e. if the data is captured using different tracking and XR systems such as Powerwall VRs combined with mobile eye tracking or HMDs with integrated eye tracking.
Another advantage is that the annotation can be done in real time or after the experiments with human test subjects was already conducted.
It is even possible to define or change the volumes of interest after the user study.
Nevertheless, our work is to be seen as a proof of concept.
We expect potential future work to further increase the accuracy of predictions.
Chances for improving our approach beyond many are advanced image classification methods or further improving the image augmentations techniques.
All in all, this paper should be seen as a contribution in solving the problem of semantic gaze analysis in three-dimensional interactive scenes.
It furthermore enables cross-platform use of eye tracking in virtual and mixed reality, a property that competing approaches lack.

\begin{acks}
Thanks to Maximilian Stubbemann for fruitful discussions, motivating words and comments on the manuscript.
We also thank our former colleague Christian Esser for the permission to use the jointly collected experimental data.
\end{acks}

\bibliographystyle{ACM-Reference-Format}
\bibliography{paper}

\appendix

\end{document}